# SYNTACTIC ANALYSIS OF NATURAL LANGUAGE USING LINGUISTIC RULES AND CORPUS-BASED PATTERNS

Pasi Tapanainen[*]

Rank Xerox Research Centre
Grenoble Laboratory

Timo Järvinen

University of Helsinki
Research Unit for Computational Linguistics

## Abstract

We are concerned with the syntactic annotation of unrestricted text. We combine a rule-based analysis with subsequent exploitation of empirical data. The rule-based surface syntactic analyser leaves some amount of ambiguity in the output that is resolved using empirical patterns. We have implemented a system for generating and applying corpus-based patterns. Some patterns describe the main constituents in the sentence and some the local context of the each syntactic function. There are several (partly) redundant patterns, and the "pattern" parser selects analysis of the sentence that matches the strictest possible pattern(s). The system is applied to an experimental corpus. We present the results and discuss possible refinements of the method from a linguistic point of view.

## 1 INTRODUCTION

We discuss surface-syntactic analysis of running text. Our purpose is to mark each word with a syntactic tag. The tags denote subjects, object, main verbs, adverbials, etc. They are listed in Appendix A.

Our method is roughly following

- Assign to each word all the possible syntactic tags.

- Disambiguate words as much as possible using linguistic information (hand-coded rules). Here we avoid risks; we rather leave words ambiguous than guess wrong.

- Use global patterns to form alternative sentence level readings. Those alternative analyses are selected that match the strictest global pattern. If it does not accept any of the remaining readings, the second strictest pattern is used, and so on.

- Use local patterns to rank the remaining readings. The local patterns contain possible contexts for syntactic functions. The ranking of the readings depends on the length of the contexts associated with the syntactic functions of the sentece.

We use both linguistic knowledge, represented as rules, and empirical data collected from tagged corpora. We describe a new way to collect information from a tagged corpus and a way to apply it. In this paper, we are mainly concerned with exploiting the empirical data and combining two different kinds of parsers.

Our work is based on work done with ENGCG, *the Constraint Grammar Parser of English* [Karlsson, 1990; Karlsson, 1994; Karlsson *et al.*, 1994; Voutilainen, 1994]. It is a rule-based tagger and surface-syntactic parser that makes a very small number of errors but leaves some words ambiguous i.e. it prefers ambiguity to guessing wrong. The morphological part-of-speech analyser leaves [Voutilainen *et al.*, 1992] only 0.3 % of all words in running text without the correct analysis when 3–6 % of words still have two or more[1] analyses.

Voutilainen, Heikkilä and Anttila [1992] reported that the syntactic analyser leaves 3–3.5 % of words without the correct syntactic tag, and 15–20 % of words remain ambiguos. Currently, the error rate has been decreased to 2–2.5 % and ambiguity rate to 15 % by Timo Järvinen [1994], who is responsible for tagging a 200 million word corpus using ENGCG in the *Bank of English* project.

Althought, the ENGCG parser works very well in part-of-speech tagging, the syntactic descriptions are still problematic. In the constraint grammar framework, it is quite hard to make linguistic generalisations that can be applied reliably. To resolve the remaining ambiguity we generate, by using a tagged corpus, a knowledge-base that contains information about both the general structure of the sentences and the local contexts of the syntactic tags. The general structure contains information about where, for example, subjects, objects and main verbs appear and how they follow one another. It does not pay any attention to their potential modifiers. The modifier–head relations are resolved by using the local context i.e. by looking at what kinds of words there are in the neighbourhood.

The method is robust in the sense that it is able to handle very large corpora. Although rule-based parsers usually perform slowly, that is not the case with ENGCG. With the English grammar, the Constraint Grammar Parser implementation by Pasi Tapanainen analyses 400 words[2] per second on a SparcStation 10/30. That is, one million words are processed in about 40 minutes. The pattern parser for empirical patterns runs somewhat slower, about 100 words per second.

---

[*]This work was done when the author worked in the Research Unit for Computational Linguistics at the University of Helsinki.

[1]But even then some of the original alternative analyses are removed

[2]Including all steps of preprocessing, morphological analysis, disambiguation and syntactic analysis. The speed of morphological disambiguation alone exceeds 1000 words per second.

## 2 KNOWLEDGE ACQUISITION

We have used two schemes to extract knowledge from corpora. Both produce readable patterns that can be verified by a linguist. In the first scheme, sentences are handled as units and information about the structure of the sentence is extracted. Only the main constituents (like subject, objects) of the sentence are treated at this stage. The second scheme works with local context and looks only a few words to the right and to the left. It is used to resolve the modifier–head dependencies in the phrases.

First, we form an *axis of the sentence* using some given set of syntactic tags. We collect several layers of patterns that may be partly redundant with each other. For instance, simplifying a little, we can say that a sentence can be of the form *subject — main verb* and there may be other words before and after the subject and main verb. We may also say that a sentence can be of the form *subject — main verb — object*. The latter is totally covered by the former because the former statement does not prohibit the appearance of an object but does not require it either.

The redundant patterns are collected on purpose. During parsing we try to find the strictest frame for the sentence. If we can not apply some pattern because it conflicts with the sentence, we may use other, possibly more general, pattern. For instance, an axis that describes all accepted combinations of subject, objects and main verbs in the sentence, is stricter than an axis that describes all accepted combinations of subjects and main verbs.

After applying the axes, the parser's output is usually still ambiguous because all syntactic tags are not taken into account yet (we do not handle, for instance, determiners and adjective premodifiers here). The remaining ambiguity is resolved using local information derived from a corpus. The second phase has a more probabilistic flavour, although no actual probabilities are computed. We represent information in a readable form, where all possible contexts, that are common enough, are listed for each syntactic tag. The length of the contexts may vary. The common contexts are longer than the rare ones. In parsing, we try to find a match for each word in a maximally long context.

Briefly, the relation between the axes and the joints is following. The axes force sentences to comply with the established frames. If more than one possibility is found, the joints are used to rank them.

### 2.1 The sentence axis

In this section we present a new method to collect information from a tagged corpus. We define a new concept, a *sentence axis*. The sentence axis is a pattern that describes the sentence structure at an appropriate level. We use it to select a group of possible analyses for the sentence. In our implementation, we form a group of sentence axes and the parser selects, using the axes, those analyses of the sentence that match all or as many as possible sentence axes.

We define the sentence axis in the following way. Let S be a set of sentences and T a set of syntactic tags. The *sentence axis* of S according to tags T shows the order of appearance of any tag in T for every sentence in S.

Here, we will demonstrate the usage of a sentence axis with one sentence. In our real application we, of course, use more text to build up a database of sentence axes. Consider the following sentence[3]

*I*_SUBJ *would*_+FAUXV *also*_ADVL
*increase*_-FMAINV *child*_NN> *benefit*_OBJ ,
*give*_-FMAINV *some*_QN> *help*_OBJ
*to*_ADVL *the*_DN> *car*_NN> *industry*_<P
*and*_CC *relax*_-FMAINV *rules*_OBJ
*governing*_<NOM-FMAINV *local*_AN>
*authority*_NN> *capital*_AN> *receipts*_OBJ ,
*allowing*_-FMAINV *councils*_SUBJ
*to*_INFMARK> *spend*_-FMAINV *more*_ADVL .

The axis according to the manually defined set T =

{ SUBJ +FAUXV +FMAINV }

is

··· SUBJ +FAUXV ··· SUBJ ···

which shows what order the elements of set T appear in the sentence above, and where three dots mean that there may be something between words, e.g. *+FAUXV* is not followed (in this case) immediately by *SUBJ*. When we have more than one sentence, the axis contains more than one possible order for the elements of set T.

The axis we have extracted is quite general. It defines the order in which the finite verbs and subjects in the sentence may occur but it does not say anything about nonfinite verbs in the sentence. Notice that the second subject is not actually the subject of the finite clause, but the subject of nonfinite construction *councils to spend more*. This is inconvenient, and a question arises whether there should be a specific tag to mark subjects of the nonfinite clauses. Voutilainen and Tapanainen [1993] argued that the richer set of tags could make parsing more accurate in a rule-based system. It may be true here as well.

We can also specify an axis for verbs of the sentence. Thus the axis according to the set

{ +FAUXV +FMAINV
-FMAINV INFMARK> }

is

··· +FAUXV ··· -FMAINV ··· -FMAINV
··· -FMAINV ··· -FMAINV ··· INFMARK>
-FMAINV ···

The nonfinite verbs occur in this axis four times one after another. We do not want just to list how many times a nonfinite verb may occur (or occurs in a corpus) in this kind of position, so we clearly need some generalisations.

The *fundamental rule of generalisation* that we used is the following: Anything that is repeated may be repeated any number of times.

We mark this using brackets and a plus sign. The generalised axis for the above axis is

··· +FAUXV [ ··· -FMAINV ]+
··· INFMARK> -FMAINV ···

---

[3]The tag set is adapted from the Constraint Grammar of English as it is. It is more extensive than commonly used in tagged corpora projects (see Appendix A).

We can also repeat longer sequences, for instance the set

{ –FMAINV <NOM–FMAINV +FAUXV SUBJ OBJ }

provides the axis

SUBJ +FAUXV ⋯ –FMAINV ⋯ OBJ
⋯ –FMAINV ⋯ OBJ ⋯ –FMAINV OBJ
⋯ <NOM–FMAINV ⋯ OBJ ⋯
–FMAINV SUBJ ⋯ –FMAINV ⋯

And we form a generalisation

SUBJ +FAUXV [ ⋯ –FMAINV ⋯ OBJ ]+
⋯ <NOM–FMAINV ⋯ OBJ ⋯
–FMAINV SUBJ ⋯ –FMAINV ⋯

Note that we added silently an extra dot between one *–FMAINV* and *OBJ* in order not to make distinctions between *–FMAINV OBJ* and *–FMAINV* ⋯ *OBJ* here.

Another generalisation can be made using equivalence classes. We can assign several syntactic tags to the same equivalence class (for instance *–FMAINV*, *<NOM-FMAINV* and *<P-FMAINV*), and then generate axes as above. The result would be

SUBJ +FAUXV [ ⋯ nonfinv ⋯ OBJ ]+
⋯ nonfinv SUBJ ⋯ nonfinv ⋯

where *nonfinv* denotes both *–FMAINV* and *<NOM-FMAINV* (and also *<P-FMAINV*).

The equivalence classes are essential in the present tag set because the syntactic arguments of finite verbs are not distinguished from the arguments of nonfinite verbs. Using equivalence classes for the finite and nonfinite verbs, we may build an generalisation that applies to both types of clauses. Another way to solve the problem, is to add new tags for the arguments of the nonfinite clauses, and make several axes for them.

### 2.2 Local patterns

In the second phase of the pattern parsing scheme we apply local patterns, the *joints*. They contain information about what kinds of modifiers have what kinds of heads, and vice versa.

For instance, in the following sentence[4] the words *fair* and *crack* are both three ways ambiguous before the axes are applied.

*He*_SUBJ *gives*_+FMAINV *us*_I-OBJ
*a*_DN> *fair*_AN>/SUBJ/NN>
*crack*_OBJ/+FMAINV/SUBJ *then*_ADVL
*we*_SUBJ *will*_+FAUXV
*be*_–FMAINV/-FAUXV *in*_ADVL *with*_ADVL
*a*_DN> *chance*_<P *of*_<NOM-OF
*carrying*_<P-FMAINV *off*_<NOM/ADVL
*the*_DN> *World*_<P/NN> *Cup*_<P/OBJ .

After the axes have been applied, the noun phrase *a fair crack* has the analyses

*a*_DN> *fair*_AN>/NN> *crack*_OBJ.

The word *fair* is still left partly ambiguous. We resolve this ambiguity using the joints.

In an ideal case we have only one head in each phrase, although it may not be in its exact location yet. The following sentence fragment demonstrates this

*They*_SUBJ *have*_+FAUXV *been*_–FMAINV
*much*_AD-A> *less*_PCOMPL-S/AD-A>
*attentive*_<NOM/PCOMPL-S
*to*_<NOM/ADVL *the*_DN> ⋯

In the analysis, the head of the phrase *much less attentive* may be *less* or *attentive*. If it is *less* the word *attentive* is a postmodifier, and if the head is *attentive* then *less* is a premodifier. The sentence is represented internally in the parser in such a way that if the axes make this distinction, i.e. force there to be exactly one subject complement, there are only two possible paths which the joints can select from: *less*_AD-A> *attentive*_PCOMPL-S and *less*_PCOMPL-S *attentive*_<NOM.

Generating the joints is quite straightforward. We produce different alternative variants for each syntactic tag and select some of them. We use a couple of parameters to validate possible *joint* candidates.

- The error margin provides the probability for checking if the context is relevant, i.e., there is enough evidence for it among the existing contexts of the tag. This probability may be used in two ways:
  - For a syntactic tag, generate all contexts (of length $n$) that appear in the corpora. Select all those contexts that are frequent enough. Do this with all $n$'s values: 1, 2, ...
  - First generate all contexts of length 1. Select those contexts that are freguent enough among the generated contexts. Next, lengthen all contexts selected in the previous step by one word. Select those contexts that are frequent enough among the new generated contexts. Repeat this sufficient many times.

  Both algorithms produce a set of contexts of different lengths. Characteristic for both the algorithms is that if they have generated a context of length $n$ that matches a syntactic function in a sentence, there is also a context of length $n-1$ that matches.

- The absolute margin – number of cases that is needed for the evidence of the generated context. If there is less evidence, it is not taken into account and a shorter context is generated. This is used to prevent strange behaviour with syntactic tags that are not very common or with a corpus that is not big enough.

- The maximum length of the context to be generated.

During the parsing, longer contexts are preferred to shorter ones. The parsing problem is thus a kind of pattern matching problem: we have to match a pattern (context) around each tag and find a sequence of syntactic tags (analysis of the sentence) that has the best score. The scoring function depends on the lengths of the matched patterns.

---
[4]This analysis is comparable to the output of ENGCG. The ambiguity is marked here using the slash. The morphological information is not printed.

| text  | words | ambiguity rate | error rate |
|-------|-------|----------------|------------|
| bb1   | 1734  | 12.4 %         | 2.4 %      |
| bb2   | 1674  | 14.2 %         | 2.8 %      |
| today | 1599  | 18.6 %         | 1.6 %      |
| wsj   | 2309  | 16.2 %         | 2.9 %      |
| total | 7316  | 15.3 %         | 2.2 %      |

Figure 1: Test corpora after syntactical analysis of ENGCG.

## 3 EXPERIMENTS WITH REAL CORPORA

Information concerning the axes was acquired from a manually checked and fully disambiguated corpus[5] of about 30,000 words and 1,300 sentences. Local context information was derived from corpora that were analysed by ENGCG. We generated three different parsers using three different corpora[6]. Each corpus contains about 10 million words.

For evaluation we used four test samples (in Figure 1). Three of them were taken from corpora that we used to generate the parsers and one is an additional sample. The samples that are named *bb1*, *today* and *wsj* belong to the corpora from which three different *joint* parsers, called *BB1*, *TODAY* and *WSJ* respectively, were generated. Sample *bb2* is the additional sample that is not used during development of the parsers.

The ambiguity rate tells us how much ambiguity is left after ENGCG analysis, i.e. how many words still have one or more alternative syntactic tags. The error rate shows us how many syntactic errors ENGCG has made while analysing the texts. Note that the ambiguity denotes the amount of work to be done, and the error rate denotes the number of errors that already exist in the input of our parser.

All the samples were analysed with each generated parser (in Figure 2). The idea is to find out about the effects of different text types on the generation of the parsers. The present method is applied to reduce the syntactic ambiguity to zero. Success rates variate from 88.5 % to 94.3 % in different samples. There is maximally a 0.5 percentage points difference in the success rate between the parsers when applied to the same data. Applying a parser to a sample from the same corpus of which it was generated does not generally show better results.

Some of the distinctions left open by ENGCG may not be structurally resolvable (see [Karlsson *et al.*, 1994]). A case in point is the prepositional attachment ambiguity, which alone represents about 20 % of the ambiguity in the ENGCG output. The proper way to deal with it in the CG framework is probably using lexical information.

Therefore, as long as there still is structurally unresolvable ambiguity in the ENGCG output, a certain amount of processing before the present system might improve the results considerably, e.g., converting structurally unresolvable syntactic tags to a single underspecified tag. For instance, resolving prepositional attachment ambiguity by other means would improve the success rate of the current system to 90.5 % – 95.5 %. In the *wsj* sample the improvement would be as much as 2.0 percentage points.

| Texts | Parsers | | |
|-------|---------|---------|---------|
|       | BB1     | TODAY   | WSJ     |
| bb1   | 94.2 %  | 94.3 %  | 94.2 %  |
| bb2   | 92.5 %  | 92.5 %  | 92.8 %  |
| today | 93.0 %  | 92.8 %  | 92.8 %  |
| wsj   | 88.5 %  | 89.0 %  | 88.9 %  |
| total | 91.7 %  | 91.9 %  | 91.9 %  |

Figure 2: Overall parsing success rate in syntactically analysed samples

The differences between success rates in different samples are partly explained by the error types that are characteristic of the samples. For example, in the *Wall Street Journal* adverbials of time are easily parsed erroneously. This may cause an accumulation effect, as happens in the following sentence

> MAN AG Tuesday said fiscal 1989 net income rose 25% and said it will raise its dividend for *the year ended June 30* by about the same percentage.

The phrase *the year ended June 30* gets the analysis

> *the*_DN> *year*_NN> *ended*_AN> *June*_NN> *30*_<P

while the correct (or wanted) result is

> *the*_DN> *year*_<P *ended*_<NOM–FMAINV *June*_ADVL *30*_<NOM

Different kind of errors appear in text *bb1* which contains incomplete sentences. The parser prefers complete sentences and produces errors in sentences like

> There was Provence in mid-autumn. *Gold tints. Air* so serene you could look out over the sea for tens of miles. *Rehabilitation walks* with him along the woodland paths.

The errors are: *gold tints* is parsed as *subject – main verb* as well as *rehabilitation walks*, and *air* is analysed as a main verb. Other words have the appropriate analyses.

The strict sequentiality of morphological and syntactic analysis in ENGCG does not allow the use of syntactic information in morphological disambiguation. The present method makes it possible to prune the remaining morphological ambiguities, i.e. do some part-of-speech tagging. Morphological ambiguity remains unresolved if the chosen syntactic tag is present in two or more morphological readings of the same word. Morphological ambiguity[7] is reduced close to zero (about 0.3 % in all the samples together) and the overall success rate of ENGCG + our pattern parser is 98.7 %.

## 4 CONCLUSION

We discussed combining a linguistic rule-based parser and a corpus-based empirical parser. We divide the parsing process into two parts: applying linguistic information and applying corpus-based patterns. The linguistic rules are regarded as more reliable than the corpus-based generalisations. They are therefore applied first.

The idea is to use reliable linguistic information as long as it is possible. After certain phase it comes harder and harder to make new linguistic constraints to eliminate the remaining ambiguity. Therefore we use corpus-based patterns to do the remaining disambiguation. The overall success rate of the combination of the linguistic rule-based parser and the corpus-based pattern parser is good. If some unresolvable ambiguity is left pending (like prepositional attachment), the total success rate of our morphological and surface-syntactic analysis is only slightly worse than that of many probabilistic part-of-speech taggers. It is a good result because we do more than just label each word with a morphological tags (i.e. noun, verb, etc.), we label them also with syntactic function tags (i.e. subject, object, subject complement, etc.).

Some improvements might be achieved by modifying the syntactic tag set of ENGCG. As discussed above, the (syntactic) tag set of the ENGCG is not probably optimal. Some ambiguity is not resolvable (like prepositional attachment) and some distinctions are not made (like subjects of the finite and the nonfinite clauses). A better tag set for surface-syntactic parsing is presented in [Voutilainen and Tapanainen, 1993]. But we have not modified the present tag set because it is not clear whether small changes would improve the result significantly when compared to the effort needed.

Although it is not possible to fully disambiguate the syntax in ENGCG, the rate of disambiguation can be improved using a more powerful linguistic rule formalism (see [Koskenniemi et al., 1992; Koskenniemi, 1990; Tapanainen, 1991]). The results reported in this study can most likely be improved by writing a syntactic grammar in the finite-state framework. The same kind of pattern parser could then be used for disambiguating the resulting analyses.

## 5 ACKNOWLEDGEMENTS

The Constraint Grammar framework was originally proposed by Fred Karlsson [1990]. The extensive work on the description of English was done by Atro Voutilainen, Juha Heikkilä and Arto Anttila [1992]. Timo Järvinen [1994] has developed the syntactic constraint system further. ENGCG uses Kimmo Koskenniemi's [1983] two-level morphological analyser and Pasi Tapanainen's implementation of Constraint Grammar parser.

We want to thank Fred Karlsson, Lauri Karttunen, Annie Zaenen, Atro Voutilainen and Gregory Grefenstette for commenting this paper.

## A THE TAG SET

This appendix contains the syntactic tags we have used. The list is adopted from [Voutilainen et al., 1992]. To obtain also the morphological "part-of-speech" tags you can send an empty e-mail message to *engcg-info@ling.helsinki.fi*.

+FAUXV = Finite Auxiliary Predicator: *He can read.*,
−FAUXV = Nonfinite Auxiliary Predicator: *He may have read.*,
+FMAINV = Finite Main Predicator: *He reads.*,
−FMAINV = Nonfinite Main Predicator: *He has read.*,
NPHR = Stray NP: *Volume I: Syntax*,
SUBJ = Subject: *He reads.*,
F–SUBJ = Formal Subject: *There was some argument about that. It is raining.*,
OBJ = Object: *He read a book.*,
I–OBJ = Indirect Object: *He gave Mary a book.*,
PCOMPL–S = Subject Complement: *He is a fool.*,
PCOMPL–O = Object Complement: *I consider him a fool.*,
ADVL = Adverbial: *He came home late. He is in the car.*,
O–ADVL = Object Adverbial: *He ran two miles.*
APP = Apposition: *Helsinki, the capital of Finland,*
N = Title: *King George and Mr. Smith,*
DN> = Determiner: *He read the book.*,
NN> = Premodifying Noun: *The car park was full.*,
AN> = Premodifying Adjective: *The blue car is mine.*,
QN> = Premodifying Quantifier: *He had two sandwiches and some coffee.*,
GN> = Premodifying Genitive: *My car and Bill's bike are blue.*,
AD-A> = Premodifying Ad-Adjective: *She is very intelligent.*,
<NOM-OF = Postmodifying Of: *Five of you will pass.*,
<NOM–FMAINV = Postmodifying Nonfinite Verb: *He has the licence to kill. John is easy to please. The man drinking coffee is my uncle.*,
<AD-A = Postmodifying Ad-Adjective: *This is good enough.*,
<NOM = Other Postmodifier: *The man with glasses is my uncle. He is the president elect. The man in the moon fell down too soon.*,
INFMARK> = Infinitive Marker: *John wants to read.*,
<P-FMAINV = Nonfinite Verb as Complement of Preposition: *This is a brush for cleaning.*,
<P = Other Complement of Preposition: *He is in the car.*,
CC = Coordinator: *John and Bill are friends.*,
CS = Subordinator: *If John is there, we shall go, too.*,